\documentclass[journal]{IEEEtran}

\usepackage{cite}

\usepackage[pdftex]{graphicx}
\graphicspath{{../imgs/}}
\usepackage{amsmath}
\usepackage{amssymb}
\usepackage{algorithmic}
\usepackage{array}

\usepackage{epsfig} 
\usepackage{times}
\usepackage{svg}
\usepackage{subcaption}
\usepackage{caption}
\usepackage{hyperref}
\DeclareMathOperator*{\argmax}{argmax} % thin space, 
\usepackage{fixltx2e}
\usepackage{booktabs} % commands to create good-looking tables
\usepackage{multirow}
\usepackage{multicol}

\newcommand{\change}[1]{\textcolor{black}{#1}}

\begin{document}

\title{Increasing the Efficiency of Policy Learning for Autonomous Vehicles by Multi-Task Representation Learning}

\author{Eshagh~Kargar,
        Ville~Kyrki
        % <-this % stops a space
        \thanks{Eshagh Kargar and Ville Kyrki are with Department of Electrical Engineering and Automation, School of Electrical Engineering, Aalto University, Finland. {firstname.lastname}@aalto.fi}%
}
% \thanks{M. Shell was with the Department
% of Electrical and Computer Engineering, Georgia Institute of Technology, Atlanta,
% GA, 30332 USA e-mail: (see http://www.michaelshell.org/contact.html).}% <-this % stops a space
% \thanks{J. Doe and J. Doe are with Anonymous University.}% <-this % stops a space
% \thanks{Manuscript received April 19, 2005; revised August 26, 2015.}}

\markboth{IEEE Transactions on Intelligent Vehicles}%
{Kargar \MakeLowercase{\textit{et al.}}: Policy Learning for Autonomous Vehicles by Multi-Task Representation Learning}

\maketitle

\begin{abstract}
Driving in a dynamic, multi-agent, and complex urban environment is a difficult task requiring a complex decision-making policy. The learning of such a policy requires a state representation that can encode the entire environment. Mid-level representations that encode a vehicle's environment as images have become a popular choice. Still, they are quite high-dimensional, limiting their use in data-hungry approaches such as reinforcement learning. In this article, we propose to learn a low-dimensional and rich latent representation of the environment by leveraging the knowledge of relevant semantic factors. To do this, we train an encoder-decoder deep neural network to predict multiple application-relevant factors such as the trajectories of other agents and the ego car. Furthermore, we propose a hazard signal based on other vehicles' future trajectories and the planned route which is used in conjunction with the learned latent representation as input to a down-stream policy. We demonstrate that using the multi-head encoder-decoder neural network results in a more informative representation than a standard single-head model. In particular, the proposed representation learning and the hazard signal help reinforcement learning to learn faster, with increased performance and less data than baseline methods.
\end{abstract}


\IEEEpeerreviewmaketitle

\section{Introduction}
% Introduction to autonomous driving.
\IEEEPARstart{D}{riving} in unstructured and dynamic urban environments is an arduous task. Many moving agents such as cars, bicycles, and pedestrians affect driver's behavior and decisions. To drive a car, a driver, whether a human being or an artificial agent, needs to perceive and understand other agents' behaviors, plans, and interactions in the environment, to react accordingly. The number of factors in the scene makes the state space of this problem very large, and safe autonomous driving in this complexity is an open challenge for the research community and industry.
% \change{test test}

\begin{figure}[ht]
\includegraphics[width=\columnwidth]{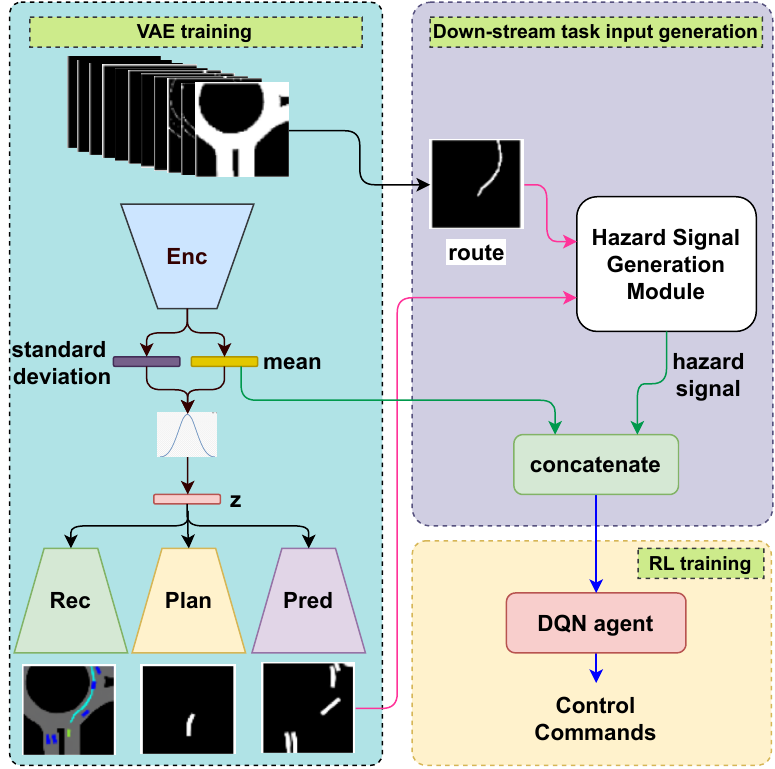}
\centering
\caption{Proposed framework. A multi-head Variational Auto-Encoder is trained using supervised learning to reconstruct a bird's eye view image of the scene, plan for ego car, and predict future trajectories of other vehicles. Route and motion prediction masks are used to calculate a hazard signal. Low-dimensional encoding of the environment and the hazard signals are used as state for reinforcement learning.}
\label{fig:prop_arch}
\end{figure}

% Representation problem, from end-to-end to mid-to-mid
A central challenge related to the high complexity of the traffic environment is representing the environment state. 

In end-to-end methods, immediate sensor measurements are used directly as the state for the decision-making policy~\cite{kendall2019learning, 9346000}. However, the  high dimensionality of the sensor data makes its direct use challenging, as a vast number of data are needed to constrain the learning problem \cite{Mller2018DrivingPT}. 
For that reason, mid-level representations, that render all perceptual inputs together with static information such as HD-map and route as bird's eye view (BEV) images, have recently received increasing attention (see e.g.~\cite{bansal2019chauffeurnet, chen2019model, kargar2021vision}). However, even the mid-level representations may be so high-dimensional that their use with data-hungry methods such as Reinforcement Learning (RL) is limited. 

% Approach
To alleviate this, we propose a new multi-task learning approach to learn low-dimensional and rich representations. 
In particular, we combine the recently proposed idea to learn a low-dimensional latent space of the mid-level image \cite{chen2019model} with the prediction of multiple auxiliary application-relevant tasks. A single latent representation is extracted from the mid-level input representations such that  the latent representation is enforced to predict the trajectories of both the ego vehicle and other vehicles in addition to a bird's eye view of the scene, using a multi-head network structure depicted in Fig.~\ref{fig:prop_arch}. All heads of the network represent the information as images, which allows their straightforward interpretation. Also, we use the motion prediction head result and the route to calculate a hazard signal as an additional input to the policy. This will give the policy information about hazardous situations and collision chance.

% Benefits
Experiments demonstrate that the auxiliary tasks allow the latent vector to learn more representative information from the scene. Therefore, a policy network can be trained faster and performs better, even with less data than a representation based on a single-head network trained for the optimal reconstruction of the scene.

% Contributions
The primary contributions of this work are: (a) a multi-task network with auxiliary heads to improve the quality of low-dimensional representations, (b) a hazard signal calculated by the likelihood between route and predicted trajectories of dynamic agents, and (c) an experimental study of an RL policy learning, showing that the learned latent-vector by using auxiliary tasks and also the hazard signal, can help the policy to be (i) trained faster, (ii) perform better, (iii) learn to solve the task using less data, \change{(iv) and generalize better to new scenarios}.

% review
The rest of this paper is organized as follows. 
The review of related works in Section \ref{related_work} demonstrates that while Reinforcement Learning has been used in autonomous driving, learning a low-dimensional state space for policy learning needs to be explored more. 
In Section \ref{background}, we provide the required background in Variational Auto-Encoder (VAE), Reinforcement Learning, and Deep Q-Network. 
The problem definition and the proposed method are described in Section \ref{method}, with emphasis on the two innovations, multi-head VAE for latent representation learning and hazard signal. 
Then, Section \ref{experiments} presents empirical evaluation in two driving scenarios showing superior performance of the proposed approach compared to state-of-the-art. 
Finally, in Section \ref{conclusion} we conclude that using task-relevant heads in VAE and also the generated hazard signal can help to learn the down-stream driving policy more efficiently.

%%%%%%%%%%%%%%%%%%%%%%%%%%%%%%%%%%%%%%%%%%%%%%%%%%%%%%%%%%%%%%%%%%%%%%%%%%%%%%%%
\section{Related Work}\label{related_work}

Reinforcement learning is a popular approach to learn a decision-making policy in autonomous driving ~\cite{chen2019model, mirchevska2018high, wang2017formulation, shi2019driving, kuutti2019end, deshpande2019deep, 9346000, kong2021enhanced, 9351818}. Despite the typically data-hungry nature of RL, Kendall et al.~\cite{kendall2019learning} recently demonstrated learning of a lane following task using real-world data gathered in a single day. However, the complexity of complex traffic environments requires great amounts of data since the state space of the decision making problem is vast. 

A solution to the high-dimensional state space is to first learn a low dimensional representation of the scene from raw sensor data. The low dimensional representation can then be used as input to a decision making policy. 
This can be achieved e.g.~by training a Variational Auto-Encoder (VAE) to form a latent representation of sensor data such as camera images \cite{bonatti2020learning}. 

Another approach to address the problem of high-dimensional state space is to use a mid-level representation, such as a bird's eye view of the current scene, as the state space. 
The mid-level representations can be constructed by engineered perception modules or by learning the mapping from sensor measurements to the mid-level representation~\cite{itkina2019dynamic}. 
Such representations are useful because they can capture the entire traffic environment around the vehicle in an interpretable fashion, but their dimensionality will still be high.
Thus, techniques that increase the amount of available data are needed to use the mid-level representations directly.
For example, Bansal et al.~\cite{bansal2019chauffeurnet} proposed to use data augmentation to learn to imitate a driving policy using bird's eye view images as input.
Similar mid-level representations can also be used for motion forecasting~\cite{cui2019multimodal, cui2019deep, chou2020predicting, phan-minh2020covernet, chai2019multipath}. 

It is also possible to combine mid-level input with learning a low-dimensional representation. 
Chen et al.~\cite{chen2019model} trained a VAE to reconstruct a bird's eye view image of the scene with rendered mid-level information, and then used the latent representation as input to an RL agent. 

To further constrain the learned latent representation to be useful, auxiliary tasks that are semantically relevant to the driving problem can be used. Hawke et al.~\cite{hawke2020urban} proposed to use auxiliary tasks in the sensor space of camera images that output segmentation, monocular depth, and optical flow to learn a better representation. The learned mid-level feature map was then used to train a policy to output control commands using imitation learning. 

Our work combines the learning of low-dimensional representations from mid-level representations with the use of auxiliary tasks to further constrain the learned representation. The proposed approach uses motion prediction of other agents as one of the auxiliary tasks, which allows us to propose a novel hazard signal that can be further used in reinforcement learning to inform of potential collisions during the learning process. 

%%%%%%%%%%%%%%%%%%%%%%%%%%%%%%%%%%%%%%%%%%%%%%%%%%%%%%%%%%%%%%%%%%%%%%%%%%%%%%%%
\section{Background}\label{background}
Next we will outline the required theoretical concepts behind the proposed method, including Variational Auto-Encoders (VAEs), Reinforcement learning, and Deep Q-Networks. 

\subsection{Variational Auto-Encoder}
Auto-Encoder (AE) is a type of artificial neural network used in data embedding and reconstruction with an encoder-decoder structure aiming to learn a compact and low-dimensional representation \(z\) for high-dimensional input data \(x\). Variational Auto-Encoder (VAE) extends AE by mapping input data to a distribution instead of a value. To learn compact representations, the latent distribution is subjected to a prior \(p(z)\), typically a normalized Gaussian distribution.  

The latent representation is then learned by minimizing the loss function~\cite{kingma2014auto}
\begin{equation}\label{vae_loss}
     L_{vae}(\phi,\theta) = - E_{q_{\theta} (z|x)} log (p_{\phi} (x|z))
     + D_{kl}(q_\theta (z|x) || p(z)) 
\end{equation}
where \(q_{\theta} (z|x)\) is the encoder network mapping inputs to latent space, \(p_{\phi} (x|z)\) is the decoder network, and \(D_{kl}\) term is the KL-divergence between the encoder output and the prior. The first term of \eqref{vae_loss} corresponds then to the reconstruction error. The second term enforces the representation to be compact. 

In order to improve disentanglement of the representation, \(\beta\)-VAE was introduced in~\cite{higgins2016beta} that instead of considering the KL-divergence between encoder and prior directly as a cost term, constrains the KL-divergence by an upper bound. Using Karush-Kuhn-Tucker conditions, the constrained optimization can be written using a Lagrangian factor \(\beta\) as an unconstrained optimization problem as 
\begin{equation}\label{beta_vae_loss}
     L_{\beta-vae}(\phi,\theta) = - E_{q_{\theta} (z|x)} log (p_{\phi} (x|z))
     + \beta D_{kl}(q_\theta (z|x) || p(z)) .
\end{equation}
When \(\beta = 1\), this corresponds to the regular VAE. Increasing \( \beta \) encourages more disentanglement with the cost of reconstruction quality.

\subsection{Reinforcement Learning}
%explain about POMDP
%We formulate the down-stream task of control policy learning as a partially observable Markov decision process (POMDP). 
Reinforcement learning aims to find an optimal control policy $\pi:S\rightarrow A$ from states to actions that maximizes total expected future rewards
%One way to learn this control policy \( \pi* \) is to use RL. The goal of an RL policy is to maximize the total future expected reward
\begin{equation}\label{rl_reward}
     R(\pi) = E_{\pi} \Bigg[ \sum_{t} \gamma^t r(s_t, a_t) \Bigg] 
\end{equation}
where \(s_t\), \(a_t\), \(r\), and \(\gamma\) are state at time \(t\), action at time \(t\), reward, and discount factor, respectively.

Value-function based RL solves the RL problem by determining an optimal value function
\(Q:S,A\rightarrow \mathbb{R}\) that describes the expected cumulative rewards when starting from a particular state and choosing a particular action
\begin{equation}
     Q^*(s, a) = \max_{\pi} E\Bigg[ \sum_{t=0} \gamma^t r(s_t, a_t) |s_0=s, a_0=a \Bigg] .
\end{equation}
Knowing the optimal value function, an optimal  policy \(\pi^*\) can then be determined as \(\pi^*(s) = \argmax\limits_{a} Q^*(s,a) \).

\subsection{Deep Q-Network}
In continuous state spaces with unknown dynamics, it is usually impossible to determine the value function exactly. 
Deep Q-Networks (DQNs)~\cite{mnih2015human} are a successful RL algorithm that use a deep neural network \(Q(s, a; \psi) \) to approximate the value function where \(\psi\) are the parameters of the neural network. 
DQN also uses a replay buffer $D=\{(s,a,r,s') \}$ to store experiences from past to expedite and stabilize learning. 
To further stabilize the learning, DQN defines a target Q-network with parameters \( \psi' \) which are updated only every \(\tau\) steps to the current \(\psi\).  
To optimize \(\psi\), the Q-learning loss
\begin{equation}
     L_{DQN}(\psi) = E_U(D)\Bigg[ \left(r+\gamma \max_{a'}Q(s',a';\psi') - Q(s,a,\psi)\right)^2 \Bigg]
\end{equation}
is minimized for a uniform sample of transitions sampled from $D$.

%%%%%%%%%%%%%%%%%%%%%%%%%%%%%%%%%%%%%%%%%%%%%%%%%%%%%%%%%%%%%%%%%%%%%%%%%%%%%%%%
\section{Method}\label{method}

%Learning a driving policy for dense urban environments directly from raw sensor data is difficult and needs a lot of data. 
The proposed method consists of two parts as illustrated in Fig.~\ref{fig:prop_arch}. First, a low-dimensional latent representation for bird's eye view images is constructed (left). Second, a driving policy is learned using the low-dimensional representation and a hazard signal that is formed from predicted future trajectories of other vehicles (right). 

In the following, we first describe the mid-level representation used as the input (Sec.~\ref{ssec:mid-level}). We then explain how the latent representation is learned using a multi-head VAE (Sec.~\ref{ssec:multi-headvae}).
We continue by defining the hazard signal in Sec.~\ref{ssec:hazard}. Finally, in Sec.~\ref{ssec:policy} we describe the policy learning.

\subsection{Mid-Level Input Representation}
\label{ssec:mid-level}
\begin{figure}[t!]
\includegraphics[width=\columnwidth]{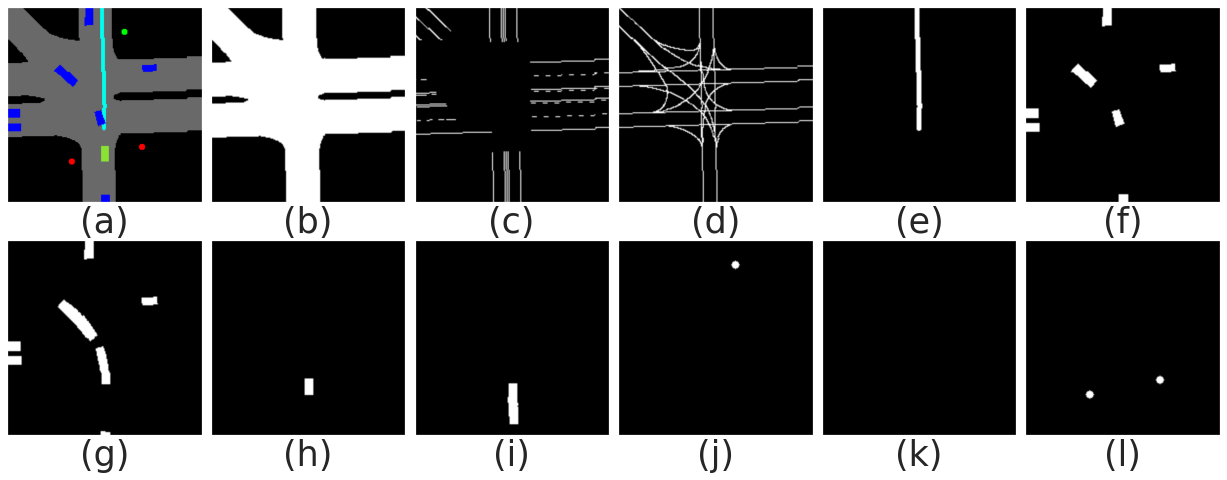}
\centering
\caption{Illustration of input channels: (a) the bird's eye view rendered RGB image, (b) road area, (c) lane lines, (d) lane centers, (e) route, (f) dynamic object's pose at the current time-step, (g) motion history of dynamic objects, (h) ego car's pose at the current time-step, (i) motion history of the ego car, (j) green traffic light, (k) yellow traffic light, and (l) red traffic light.}
\label{fig:inputs}
\end{figure}

Information in the mid-level input representation includes road relevant structures, planned route, current and past poses of ego vehicle and other vehicles, and traffic light state. These mid-level inputs are selected based on BEV input representation used in the previous works~\cite{bansal2019chauffeurnet}. Each of these is described using one or more channels of a 11-channel image, as illustrated in Fig.~\ref{fig:inputs}. 

In simulation experiments, the perception and route information is provided by the CARLA simulator \cite{dosovitskiy2017carla}. 

The information is represented as follows:
\begin{enumerate}
\item \textbf{Map}: The HD-Map information includes drivable areas, lane lines, center of lane lines, and curbs, as shown in Fig.~\ref{fig:inputs}(b--d).
\item \textbf{Route}: The route from a starting point to the current destination is assumed to originate from an external planning system such as a standard vehicle navigator. An example is shown in Fig.~\ref{fig:inputs}e.
\item \textbf{Current and past poses of other vehicles}: The current and past position, orientation, and size of other cars are rendered on two different channels, presented in Fig.\ref{fig:inputs}(f--g). We consider \(1.5s\) of motion history for dynamic targets in the scene. This information can come from the perception module with object detection and tracking algorithms and can produce other cars' current and past poses input channels by rendering these data on a single channel image. Note that the data from previous time-steps needs to be transferred to the coordinate frame of the ego car in the current time step.  
\item \textbf{Current and past ego vehicle poses}: The current and past position, orientation, and size of the ego vehicle are rendered on two different channels, described in Fig.\ref{fig:inputs}(h--i). We consider \(1.5s\) of motion history for ego vehicle. This information can come from the Localization module.
\item \textbf{Traffic lights}: The information for different traffic light colors are rendered on three different channels for green, red, and yellow as shown in Fig.\ref{fig:inputs}(j--l). Traffic lights' locations are usually available in the HD-Map, and their color can be detected using the perception module.
\end{enumerate}

BEV dimensions are 25m on the left and right side, 37.5m in front, and 12.5m behind the ego car.

\subsection{Learning Latent Representation Using Multi-Head VAE}
\label{ssec:multi-headvae}

To constrain the learning of the latent representation, it is enforced to learn three task-relevant factors: scene reconstruction, ego vehicle plan, and predicted motion of other vehicles. To do this, we use an encoder-decoder VAE which takes the 11-channel scene as the input to the encoder $q_\theta(z|x)$. The encoder transforms its input into a low-dimensional latent distribution $z$. The latent distribution is decoded into the three factors such that each factor is a separate decoder head of the neural network. The architecture for the multi-head VAE is illustrated in Fig.~\ref{fig:prop_arch} in the ``VAE training'' box.

The three decoder networks are:

\begin{enumerate}
\item \textbf{Scene Reconstruction Head} gets the latent vector \(z\) and reconstructs a three-channel bird's eye view RGB image of the scene $rgb$, rendered based on the 11-channel input similar to e.g.~\cite{chen2019model}. The neural network for this head is expressed by \( p_{\phi_1} (rgb|z) \).
\item \textbf{Planning Head} maps the latent vector to a single-channel bird's eye view planning mask for the ego vehicle for the next \(2s \). This head is represented by \( p_{\phi_2} (plan|z) \). 
\item \textbf{Motion Prediction Head} forecasts the next \(2s\) pose of dynamic agents in the scene. It gets the latent-vector \(z\) and outputs a bird's eye view mask of other agents' future motion. It is expressed by \( p_{\phi_3} (pred|z) \).
\end{enumerate}

The network architecture for each of the decoder heads is the same, but the network weights \(\phi_1, \phi_2, \phi_3 \) are different. The multi-head VAE is optimized by minimizing the loss function
\begin{equation}\label{mh_vae_loss}
    \begin{split}
     L(\theta, \phi_1,\phi_2,\phi_3)  =& - w_1 E_{q_{\theta} (z|x)} log (p_{\phi_1} (rgb|z))\\
    & - w_2 E_{q_\theta (z|x)} log (p_{\phi_2} (plan|z))  \\
     & - w_3 E_{q_\theta (z|x)} log (p_{\phi_3} (pred|z)) \\
     & + w_4 D_{kl}(q_\theta (z|x) || p(z)) 
     \end{split}
\end{equation}
with \(p(z)\) a normal distribution prior. The optimization is performed by gradient descent using a dataset of known trajectories in order to be able to train the planning and prediction heads.

\subsection{Generation of Hazard Signal}
\label{ssec:hazard}
To quantify the degree of conflict between ego car's and other vehicles' trajectories, we calculate a hazard signal $h$ as the log-likelihood that the predicted motion of other vehicles and known planned route are equal under Gaussian noise. 
% MATH NEEDED HERE!
\begin{equation}\label{hazard}
    %\begin{split}
    h = \log P(route - pred = 0) = \sum_i \log P(route_i - pred_i = 0)
    %\end{split}
\end{equation}
where the difference has Gaussian distribution, $(route_i - pred_i) \sim N(0, 1)$ for each pixel $i$ independently. 
By substituting the normal distribution density function to \eqref{hazard}, we get
\begin{equation}\label{hazard2}
h = - \sum_i \frac{(route_i - pred_i)^2}{2} + c.
\end{equation}
Thus, in the end the hazard signal is the sum of squared difference between pixel values between own route and predicted routes of other agents represented as images.

The calculated hazard signal is used in addition to the latent encoding as input to the RL policy as shown in Fig.~\ref{fig:prop_arch}.

\subsection{Policy Learning Using DQN}
\label{ssec:policy}
In order to evaluate the learned latent space, a DQN policy learning is considered as a down stream task. The DQN policy can be replaced with any other policy learning method. The latent encoding of the current state and the current hazard signal value form the input space to a DQN agent such that $s=(\mu_z,h)$, with $\mu_z$ representing the mean of the current latent encoding (see Fig.~\ref{fig:prop_arch}). The action space is a vector of three values for throttle, brake, and steering angle, $a=(a_t, a_b, a_s)$, such that each dimension is discretized into discrete choices.

The reward function is a sum of terms related to collisions, performance, obeying traffic rules (staying in lane, obeying traffic lights), and comfort:
\begin{equation}\label{eq:reward}
\begin{split}
    r = r_c + r_v + r_o +
    r_{\alpha} + r_w + r_{tl} + c
\end{split}
\end{equation}
where $r_c$ is a collision penalty, $r_v$ is a speed reward term to match desired speed, $r_o$ is an out-of-lane penalty, 
$r_{\alpha}$ is the steering angle penalty to improve driving comfort,
$r_w$ is penalty for high lateral acceleration, 
$r_{tl}$ is the penalty term for passing red traffic lights, and $c$ is a constant time penalty. Section~\ref{implementation_details} reveals more details about the weight for each reward term.
The policy is then learned using standard DQN.

%%%%%%%%%%%%%%%%%%%%%%%%%%%%%%%%%%%%%%%%%%%%%%%%%%%%%%%%%%%%%%%%%%%%%%%%%%%%%%%%
\section{Experiments}\label{experiments}
% compare the proposed method with baseline method.
We performed simulation experiments to study the following questions:
\begin{enumerate}
    \item How much is the effect of auxiliary heads and the generated hazard signal? \label{aux_heads}
    \item How our method with auxiliary heads and the hazard signal performs compared to baselines? \label{comp_baselines} 
    \item Can we decrease the dataset size by using auxiliary heads? \label{dataset_size}
    \item Can latent space learn about the scene structure, future potential trajectories for the ego agent and other agents? \label{qual_analysis}
    \item \change{How is the generalization capability of the proposed method compared to other methods?} \label{gnrlzn}
\end{enumerate}

In the following sections, first, the simulation environment and data collection have been described.
Then, the implementation details of the multi-head VAE and policy network architectures and also the reward function are detailed.
After that, the effect of different heads in the multi-head VAE is evaluated to answer question \ref{aux_heads}. The proposed method is then compared with state-of-the-art methods to answer question \ref{comp_baselines}. Dataset size effect is evaluated next to answer question \ref{dataset_size}. Then, a qualitative analysis of the learned latent-vector is done to answer question \ref{qual_analysis}. \change{Finally, to answer question \ref{gnrlzn}, the generalization capability of the proposed approach is evaluated}.
All experiments were repeated for a set of five identical random seeds for each case.

\subsection{Simulation Environment and Data Collection} \label{simenv}
\begin{figure}
     \centering
     \begin{subfigure}[b]{0.49\columnwidth}
         \centering
         \includegraphics[width=0.8\columnwidth]{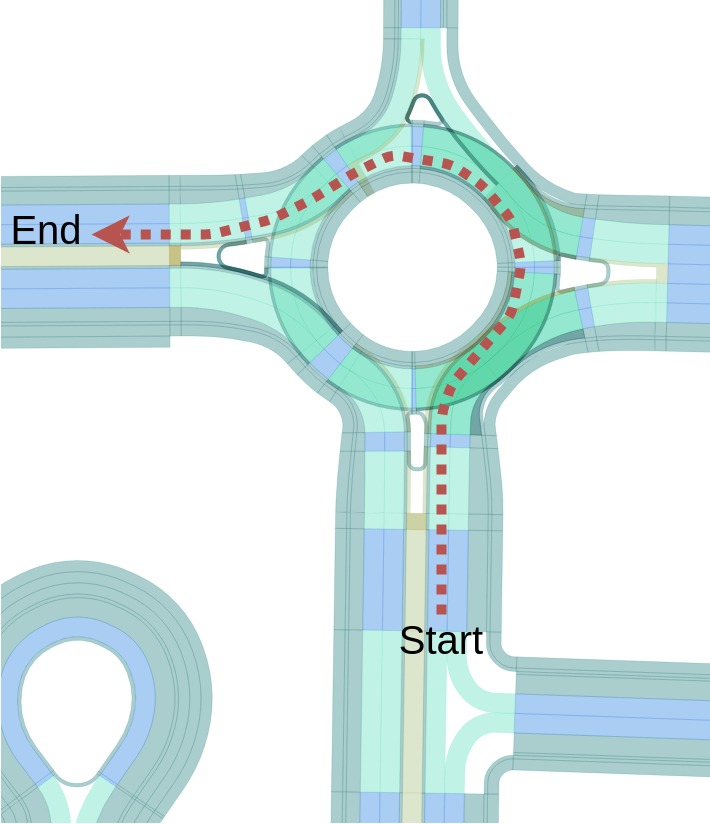}
         \caption{}
        %  \label{fig:y equals x}
     \end{subfigure}
     \hfill
     \begin{subfigure}[b]{0.49\columnwidth}
         \centering
         \includegraphics[width=0.8\columnwidth]{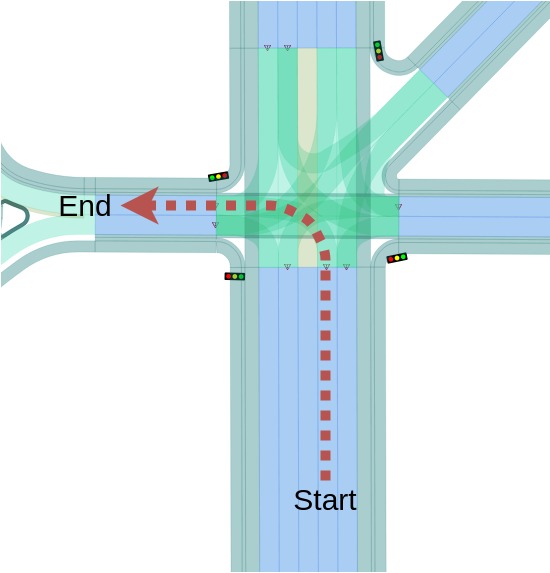}
         \caption{}
        %  \label{fig:five over x}
     \end{subfigure}
        \caption{Scenarios: (a) roundabout, (b) 5-way intersection.}
        \label{fig:scenarios}
\end{figure}

% Explain about data and simulator
To collect the dataset to train the multi-head VAE, we used the CARLA simulator~\cite{dosovitskiy2017carla}, an open-source simulator for autonomous driving, to collect the dataset. We designed two scenarios to collect the dataset: a roundabout and a complex intersection shown in Fig.~\ref{fig:scenarios}. There are no traffic lights at the roundabout, but there are several at the intersection.
For data collection phase for VAE training, we start from a random position around the start positions shown in the figure and select the destination randomly in the city. But for RL training phase, to add stochasticity, start and end positions are considered randomly around the positions shown in the figure.

For the data collection phase, 100 vehicles were spawned randomly in Town number three in CARLA and near the designed scenarios. We used CARLA autopilot mode to drive and collect a dataset. 

By recording all agents' poses in each time-step during driving, we had the required information to create the dataset without any manual labeling. To create the ground truth data for motion prediction and planning heads, we used the ego car's pose and other agents in the next time-steps, transformed them into the current ego vehicle's coordinate frame, and rendered them on a binary image. So it did not require manual labeling and was done in a self-supervised manner. The dataset size is around 200k frames.

For the RL policy learning phase, we used curriculum learning and increased number of agents from zero to 100 to increase the difficulty level step-by-step. We also set 20 percent of cars to ignore traffic lights and have aggressive driving behavior in both data collection for multi-head VAE training and RL policy training. Therefore, there is a need for multi-agent interaction and attention to other cars' movements in the intersection scenario.

% \subsection{Multi-Head VAE and Policy Network Architectures}
\subsection{Implementation Details}\label{implementation_details}
Fig.~\ref{fig:prop_arch} shows the multi-head VAE architecture in the "VAE training" box. The encoder was a ResNet-18~\cite{he2016deep} in which the first Convolutional layer had changed to get a BEV 64x64x11 input tensor. The output feature vector of the ResNet network had the size of 512. Then we used two fully connected layers with 20 neurons to output $\mu$ and $\sigma$ vectors for a Gaussian distribution on latent-space. Then the latent-vector was sampled from this distribution. There were three decoder heads with similar architecture but different weights. We used a network similar to ResNet-18 but in inverse order for decoder heads. The output of planning and prediction heads had one channel, but the reconstruction head had three channels for the output RGB image.

The weights used for the loss terms were \(w_1=1, w_2=1, w_3=50\), and \(w_4=50\) and were found using grid search.  

% \subsection{Policy Network Architecture}
The DQN policy, which is shown in the "RL training" box in Fig.~\ref{fig:prop_arch}, is a network with three fully-connected layers with 128, 64, and nine (output) neurons that gets the input vector with the size of 21 and generate the control commands. Output control commands of the policy network are: (1) acceleration $\in \{-0.3, 0, 0.3\}$ which positive value is for throttle, negative value for brake, and zero for braking and throttle, and (2) steering \(\in \{-0.15, 0, 0.15\}\).

The terms of the reward function \eqref{eq:reward} were set as follows: 
The collision penalty $r_c$ was set to $r_c = -200$ if there is a collision, otherwise $r_c = 0$. 
The speed reward $r_v$ was set to the ego vehicle's speed, $10 m/s$, and if $r_v > 10 m/s$ then a penalty of $-10$ will be added to the reward function.
The going out of lane penalty $r_o$ was set to $r_o = -1$ if the distance between the ego vehicle's position and the planned route is more than $2.5m$, otherwise $r_o = 0$.
The steering angle penalty $r_{\alpha}$ was set to $r_{\alpha} = -0.5 \times \alpha^2$.
The high lateral acceleration penalty $r_w$ was set to $r_w = -0.2 \times w$, where $w$ is the lateral acceleration.
The passing red traffic lights penalty term $r_{tl}$ (only in the intersection scenario) was set to $r_{tl} = -10$ if the ego vehicle passes a red traffic light, otherwise $r_{tl} = 0$. 
Finally, the constant time penalty $c$ was set to $-0.1$ to prevent the car from stopping.

\subsection{Effect of Different Heads and the Hazard Signal}
To answer question \ref{aux_heads} and see the effect of each head on the learned latent-vector and the proposed hazard signal, a down-stream RL task, DQN agent, was considered.
The VAE with a reconstruction head proposed in~\cite{chen2019model} was considered as the baseline (dqn\_rec). Four other models are trained to analyze the effect of each head and the hazard signal: (a) VAE with reconstruction and planning heads (dqn\_rec\_plan), (b) VAE with reconstruction and motion prediction heads (dqn\_rec\_pred), (c) VAE with reconstruction, planning, and motion prediction heads (dqn\_rec\_plan\_pred), and (d) VAE with auxiliary hazard signal in addition to three heads in case \(c\) (dqn\_rec\_plan\_pred\_hzrd).

All networks are trained using the full dataset. Then the encoder of each case was used as the feature extractor for the DQN agent. The encoder weights were fixed, and we only trained DQN weights in this phase. For case \(d\), the DQN agent also used the hazard signal. Fig.~\ref{fig:prop_arch} shows the full model used in case \(d\).

\begin{figure}
     \centering
     \begin{subfigure}[ht]{\columnwidth}
         \centering
         \includegraphics[width=0.9\columnwidth]{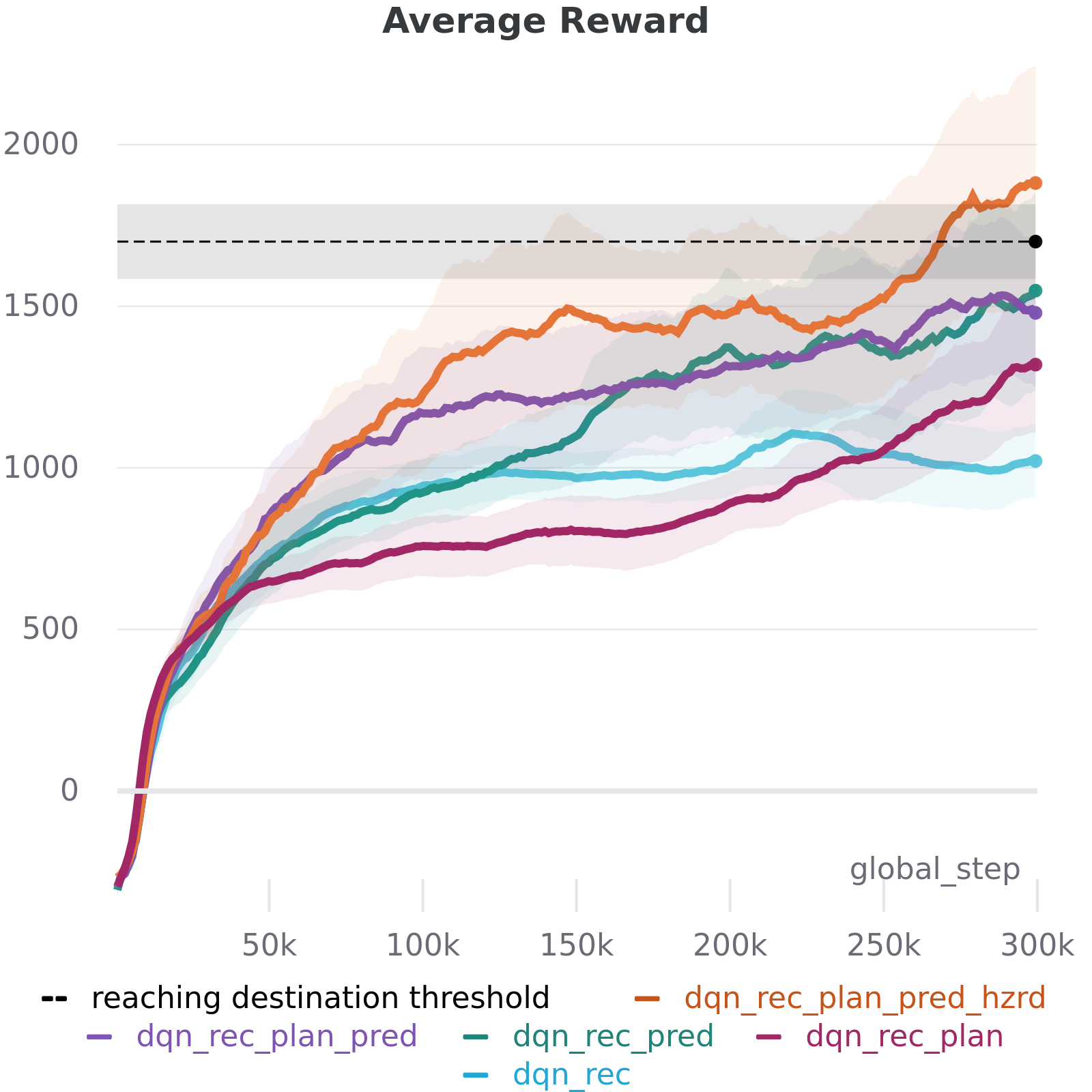}
         \caption{}
        %  \label{fig:y equals x}
     \end{subfigure}
     \hfill
     \begin{subfigure}[ht]{\columnwidth}
         \centering
         \includegraphics[width=0.9\columnwidth]{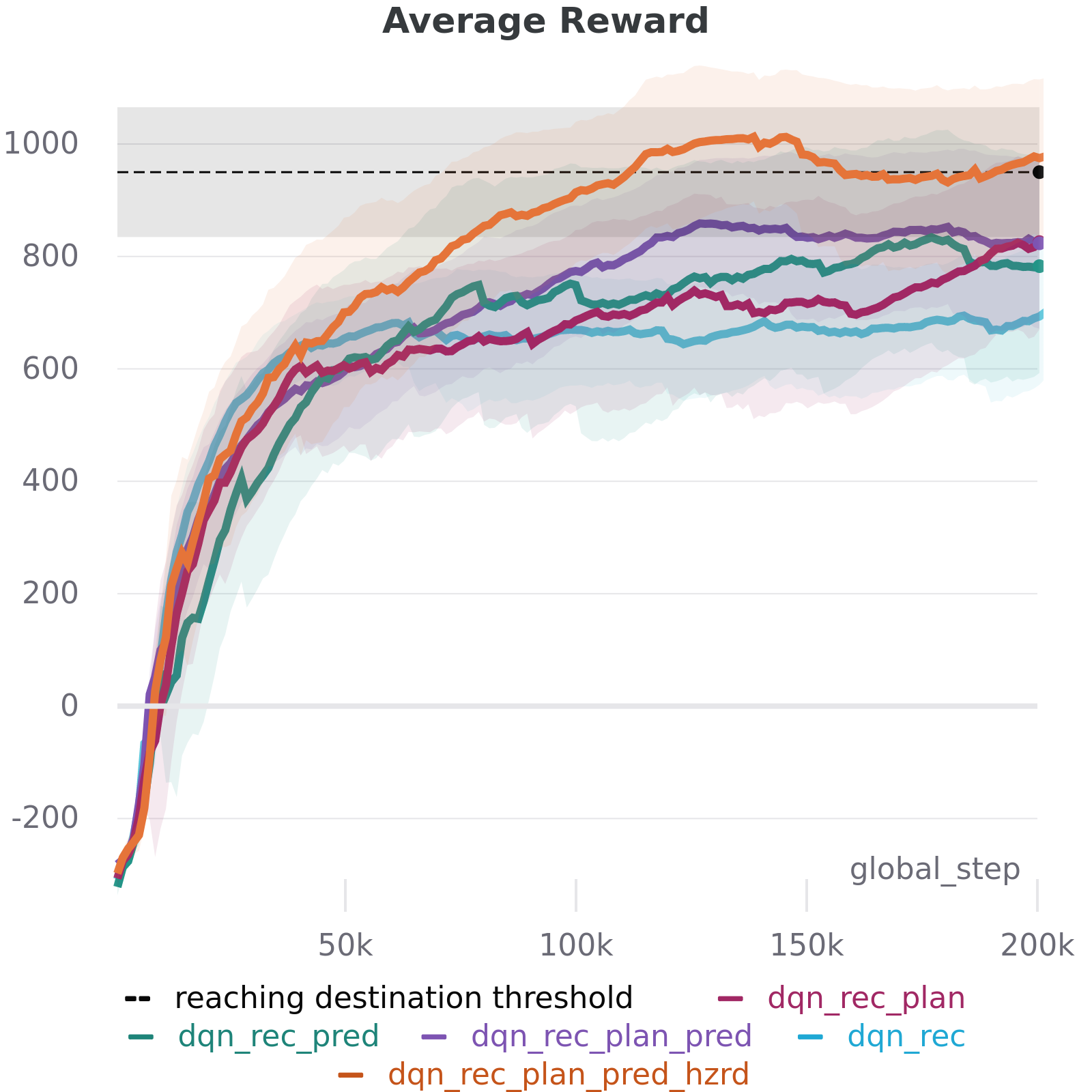}
         \caption{}
        %  \label{fig:five over x}
     \end{subfigure}
        \caption{Analysis of the effect of different VAE decoder heads and the hazard signal on the down-stream RL task's performance in two scenarios: (a) roundabout, (b) 5-way intersection.}
        \label{fig:heads}
\end{figure}

Fig.~\ref{fig:heads} shows the comparison. Our proposed method (dqn\_rec\_plan\_pred\_hzrd) outperformed all other cases, achieved a higher mean reward, and solved the task in both roundabout and intersection scenarios. The baseline method (dqn\_rec) cannot solve the task, but by adding the planning head, the agent can finally solve the task in some episodes and reach the destination. We can also see the effect of planning and prediction heads by comparing the VAE with reconstruction and prediction heads (dqn\_rec\_pred) against the VAE with reconstruction and planning heads (dqn\_rec\_plan). It shows that the motion prediction head that has information about other dynamic agents' future motion can be more beneficial than the planning head and improve the down-stream task's performance by a large margin. We can also see that using both planning and prediction heads in addition to the reconstruction head (dqn\_rec\_plan\_pred) can outperform using planning and prediction heads separately. Note that reaching the destination threshold shown in the charts isn't an exact value, but rather an area resulting from several runs of the full method (dqn\_rec\_plan\_pred\_hzrd) with different random seeds.

\begin{figure}
     \centering
     \begin{subfigure}[b]{\columnwidth}
         \centering
         \includegraphics[width=0.9\columnwidth]{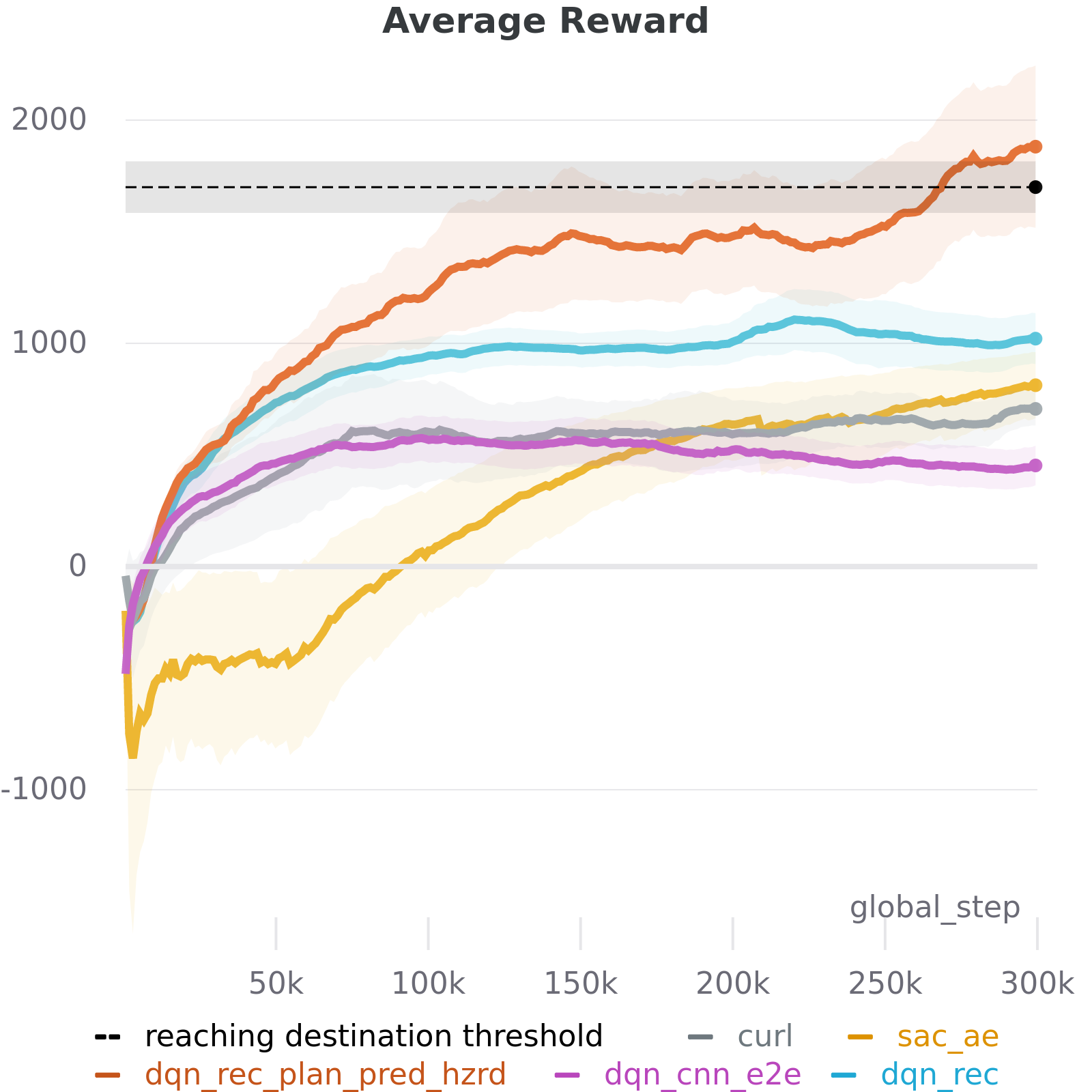}
         \caption{}
        %  \label{fig:y equals x}
     \end{subfigure}
     \hfill
     \begin{subfigure}[b]{\columnwidth}
         \centering
         \includegraphics[width=0.9\columnwidth]{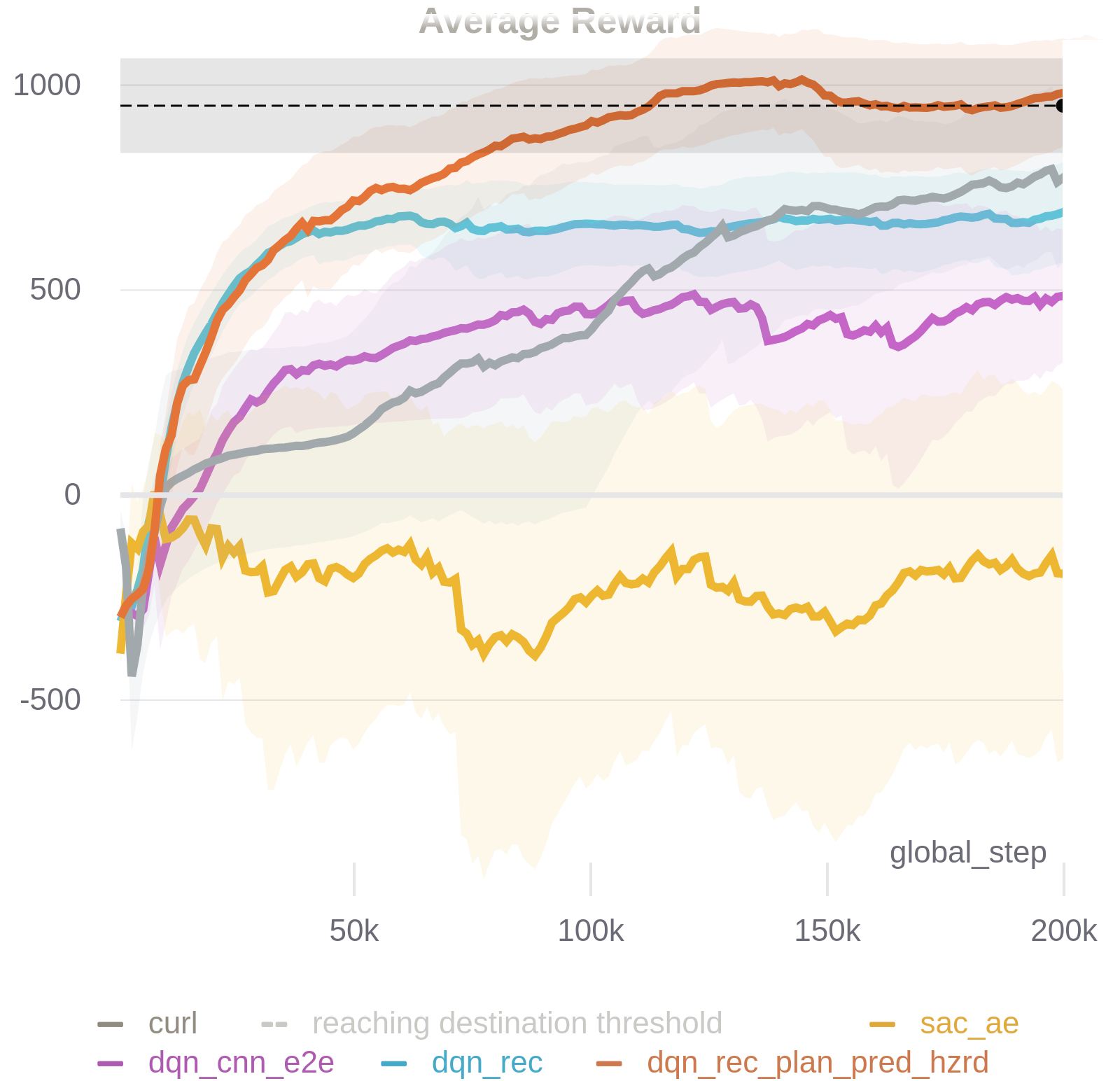}
         \caption{}
        %  \label{fig:five over x}
     \end{subfigure}
        \caption{Comparison of the proposed method against baselines in two scenarios: (a) roundabout, (b) 5-way intersection.}
        \label{fig:sota}
\end{figure}

\subsection{Comparison to Baselines}
This section compares our method with four baseline approaches in order to answer question \ref{comp_baselines}. The proposed multi-head VAE with three decoder heads and also the calculated auxiliary hazard signal as input to the DQN agent (dqn\_rec\_plan\_pred\_hzrd), shown in Fig.~\ref{fig:prop_arch}, was compared with: 
(a) an encoder and a policy network which are trained end-to-end using RL, similar to the DQN work on Atari games~\cite{mnih2013playing}, to get the BEV input tensor and generate control commands (dqn\_cnn\_e2e), 
(b) a single-head VAE with a reconstruction head (dqn\_rec)~\cite{chen2019model} which was pre-trained using supervised learning and then the weights were fixed, and the encoder was used as the feature extractor for BEV input tensor, 
(c) SAC-AE (sac\_ae)~\cite{yarats2019improving}, a variant of the standard SAC, which is more sample efficient than standard SAC and performs better in environments with image inputs,
and (d) CURL (curl)~\cite{laskin2020curl} which uses a contrastive unsupervised representation learning approach for RL algorithms.
Then the DQN agent used the encoded latent-vector as input to generate control commands.

Fig.~\ref{fig:sota} shows the comparison in two scenarios in terms of average reward. As can be seen, our proposed method (dqn\_rec\_plan\_pred\_hzrd) could solve both tasks and exceeded other methods by a considerable margin. 
Following our method, none of the baselines could solve the roundabout scenario, and VAE with reconstruction head (dqn\_rec) performed better than other baselines.
In the intersection environment, CURL (curl) worked very well and was able to accomplish the task in some runs and outperformed other methods. The next best was the VAE with reconstruction head (dqn\_rec). The SAC-AE (sac\_ae) also performed poorly in this environment. 

We ran the trained models in two different scenarios to compare different methods in terms of crash percentage and success rate. The reported results in~\cite{chen2019model} for DDQN~\cite{van2016deep} algorithm show $0\%$ success rate of reaching the goal position in the roundabout scenario and we found the same results for the dqn\_rec method. However, our method performed better and reached the success rate of $5 \pm 2 \%$ in three different runs, each consisting of 100 episodes. But it was not sufficient to get a good understanding of the performance differences. Therefore, instead of using DQN for both lateral and longitudinal control, we decided to use DQN for speed control only and use a PID controller for steering control. We only did this for the results reported in the Tables~\ref{tab:crash}, \ref{tab:success}, and \ref{tab:passtl}, not for the charts. Also, note that the reported numbers for crash and success in each case do not necessarily add up to $100\%$ and the ego car also can reach the maximum episode time. 
Additionally, we set $50 \%$ of cars to ignore traffic lights in the 5-way intersection scenario to make a more complex interactive scenario.

Tables~\ref{tab:crash}, \ref{tab:success} show the crash percentage and success rate for different methods in two scenarios for different numbers of spawned cars in the city, respectively.  
When we reduced the number of spawned cars, all methods performed better. In both scenarios, our method had a lower crash percentage and a higher success rate. 
Following that, in the roundabout scenario, SAC-AE (sac\_ae) and then VAE with reconstruction head (dqn\_rec) performed better than the others.

After our method, SAC-AE (sac\_ae), CURL (curl), and VAE with reconstruction head (dqn\_rec) all performed almost identically in the 5-way intersection scenario. This emphasizes the importance of motion prediction and hazard signals that assist ego cars to handle interactions with aggressive vehicles at an intersection.
Note that the episode was not terminated when the car passed a red traffic light in the intersection scenario, and the reported success rate numbers indicate the car reached the goal position even when it passed a red traffic light.

\begin{table}[t!]
    \centering
  \caption{The crash percentage in 3 runs, 100 episodes each, for two scenarios: Roundabout and 5-way intersection.}
  \label{tab:crash}
  \begin{tabular}{|*{5}{c|}}\toprule
    \multirow{2}{*}{\textbf{\textit{Scenario}}} & \multirow{2}{*}{\textbf{\textit{Method}}} & \multicolumn{3}{c|}{\textbf{\textit{Number of Cars}}} \\ 
     &  & \textit{10}     & \textit{50}     & \textit{100}    \\ \midrule
    
    \multirow{5}{*}{\textbf{\textit{Roundabout}}}
    
    & dqn\_cnn\_e2e  & 33 $\pm$ 3.5         & 48 $\pm$ 4               & 50 $\pm$ 3.2              \\
    & sac\_ae        & 25 $\pm$ 2.16        & 28 $\pm$ 3.3             & 31.7 $\pm$ 2.5            \\
    & curl           & 29.4 $\pm$ 1.7       & 42 $\pm$ 1.7             & 47 $\pm$ 3                \\
    & dqn\_rec       & 25 $\pm$ 4.2         & 38 $\pm$ 3.8             & 41 $\pm$ 4.2              \\
    & ours           & \textbf{6 $\pm$ 3.4} & \textbf{15.6 $\pm$ 4.1}  & \textbf{18.6 $\pm$ 5.4}   \\ \bottomrule
    
    \multirow{5}{*}{ \begin{tabular}{@{}c@{}} \textbf{\textit{5-way}} \\ \textbf{\textit{Intersection}} \end{tabular} }
    
    & dqn\_cnn\_e2e  & 33.7 $\pm$ 2.5           & 39.7 $\pm$ 2.5            & 47 $\pm$ 1.7              \\
    & sac\_ae        & 27.7 $\pm$ 1.3           & 35.7 $\pm$ 0.5            & 39.7 $\pm$ 1.3            \\
    & curl           & 25.7 $\pm$ 1.3           & 35.4 $\pm$ 1              & 41.7 $\pm$ 1.3            \\
    & dqn\_rec       & 26.7 $\pm$ 0.5           & 33.7 $\pm$ 1.7            & 40 $\pm$ 1                \\
    & ours           & \textbf{13 $\pm$ 2.2}    & \textbf{20.7 $\pm$ 1.7}   & \textbf{22.4 $\pm$ 2.1}   \\ \bottomrule
    
  \end{tabular}
\end{table}

\begin{table}[t!]
    \centering
  \caption{The Success rate in 3 runs, 100 episodes each, for two scenarios: Roundabout and 5-way intersection.}
  \label{tab:success}
  \begin{tabular}{|*{5}{c|}}\toprule
    \multirow{2}{*}{\textbf{\textit{Scenario}}} & \multirow{2}{*}{\textbf{\textit{Method}}} & \multicolumn{3}{c|}{\textbf{\textit{Number of Cars}}} \\ 
     &  & \textit{10}     & \textit{50}     & \textit{100}    \\ \midrule
    
    \multirow{5}{*}{\textbf{\textit{Roundabout}}}
    
    & dqn\_cnn\_e2e  & 59 $\pm$ 4            & 49 $\pm$ 3.1             & 47 $\pm$ 2.4              \\
    & sac\_ae        & 73.7 $\pm$ 1.7        & 71 $\pm$ 3               & 66.7 $\pm$ 1.7            \\
    & curl           & 60.4 $\pm$ 2.9        & 45 $\pm$ 1.7             & 48 $\pm$ 3                \\
    & dqn\_rec       & 65 $\pm$ 4.4          & 54.7 $\pm$ 3.2           & 52 $\pm$ 3.6              \\
    & ours           & \textbf{91 $\pm$ 3.3} & \textbf{82.4 $\pm$ 3.5}  & \textbf{79.7 $\pm$ 4.8}   \\ \bottomrule
    
    \multirow{5}{*}{ \begin{tabular}{@{}c@{}} \textbf{\textit{5-way}} \\ \textbf{\textit{Intersection}} \end{tabular} }
    
    & dqn\_cnn\_e2e  & 66.7 $\pm$ 1.3           & 59 $\pm$ 2.5           & 52.7 $\pm$ 2.1            \\
    & sac\_ae        & 70.7 $\pm$ 1.3           & 64.7 $\pm$ 1.3         & 58.7 $\pm$ 1              \\
    & curl           & 72.3 $\pm$ 1             & 63 $\pm$ 0.9           & 57.6 $\pm$ 1.7            \\
    & dqn\_rec       & 71.3 $\pm$ 1.3           & 65.4 $\pm$ 0.5         & 59.4 $\pm$ 1.3            \\
    & ours           & \textbf{84.7 $\pm$ 3.4}  & \textbf{78 $\pm$ 1.7}  & \textbf{75.7 $\pm$ 3.1}   \\ \bottomrule
    
  \end{tabular}
\end{table}

To compare the performance of methods in terms of respecting traffic lights, we evaluated them in the 5-way intersection scenario. The results are detailed in Table~\ref{tab:passtl}.
As we didn't see much difference in the performance of different algorithms with a lower number of spawned cars, we only reported the case for 100 spawned cars in the city. 

Our method outperformed others by a large margin, as shown in Table~\ref{tab:passtl}. This is because of the motion planning head, which has to imitate the driving experience of expert drivers in order to learn the correct driving response in this situation, red traffic light.

\begin{table}[t!]
    \centering
  \caption{The rate of passing traffic light in 3 runs, 100 episodes each, for the 5-way intersection scenario when 100 cars are spawned in the city.}
  \label{tab:passtl}
  \begin{tabular}{|*{5}{c|}}\toprule
     \multicolumn{5}{|c|}{\textbf{\textit{Method}}} \\ \midrule
     dqn\_cnn\_e2e & sac\_ae & curl  & dqn\_rec  & ours \\ \midrule
    50 $\pm$ 5 & 31.4 $\pm$ 2 & 34.4 $\pm$ 1.3 & 38.4 $\pm$ 1.7 &  \textbf{21 $\pm$ 2.5}   \\ \bottomrule
  \end{tabular}
\end{table}

\subsection{Effect of the Dataset Size}
In this part we try to investigate question \ref{dataset_size}. Several models were trained using different dataset sizes to show the benefit of using multiple decoder heads and the hazard signal on the learned latent-vector and the down-stream task's performance.
The full model, shown in Fig.~\ref{fig:prop_arch}, was trained using the full dataset (dqn\_rec\_plan\_pred\_hzrd\_full), half of the dataset (dqn\_rec\_plan\_pred\_hzrd\_half), and a quarter of the dataset (dqn\_rec\_plan\_pred\_hzrd\_quart). We then compared them against the baseline VAE~\cite{chen2019model} with just a reconstruction decoder (dqn\_rec) trained on the full dataset.
For our proposed method, a DQN agent was trained to get the concatenation of the latent-vector and the hazard signal and output control commands. Note that for the baseline, there is no hazard signal, as there is no prediction head.

\begin{figure}
     \centering
     \begin{subfigure}[b]{\columnwidth}
         \centering
         \includegraphics[width=0.9\columnwidth]{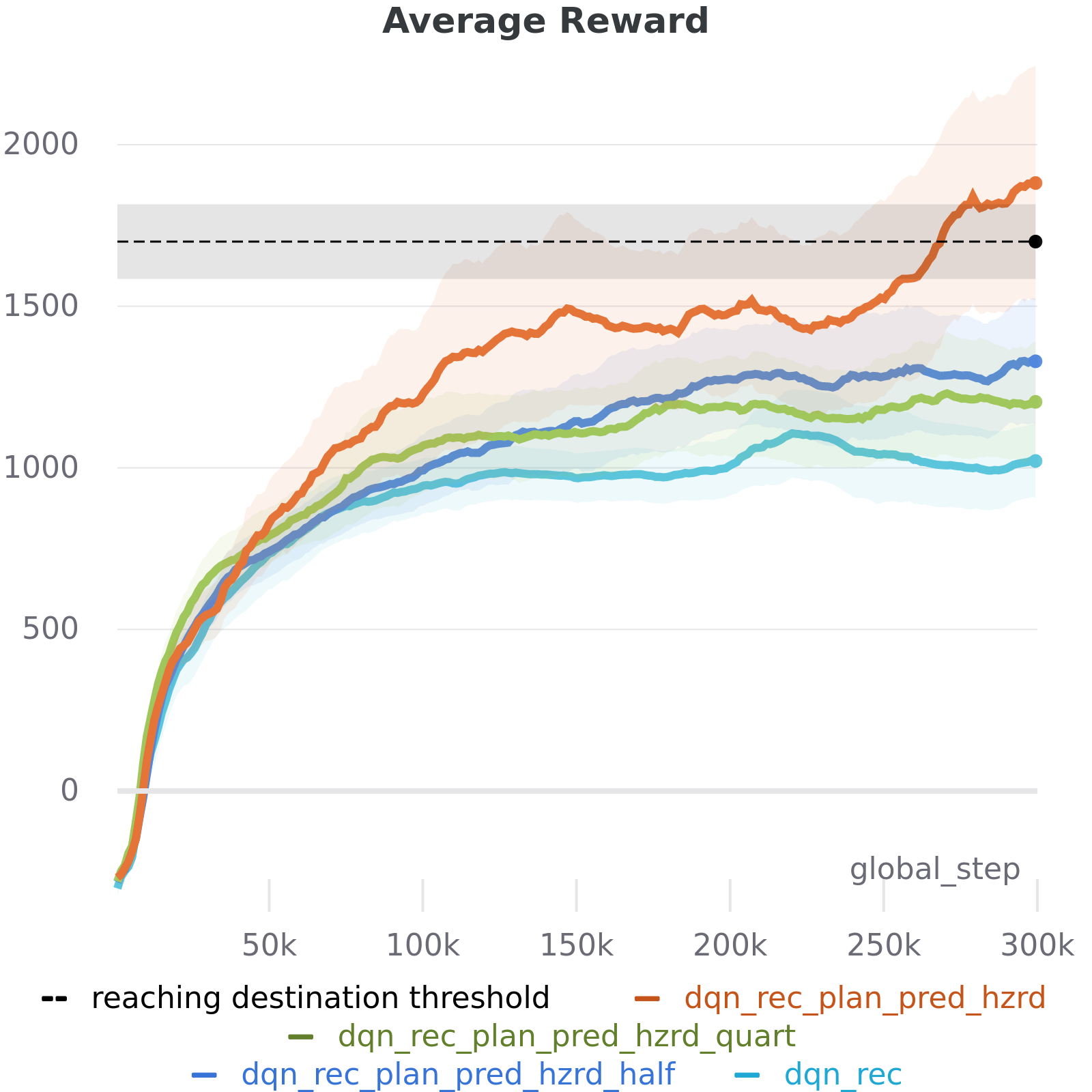}
         \caption{}
        %  \label{fig:y equals x}
     \end{subfigure}
     \hfill
     \begin{subfigure}[b]{\columnwidth}
         \centering
         \includegraphics[width=0.9\columnwidth]{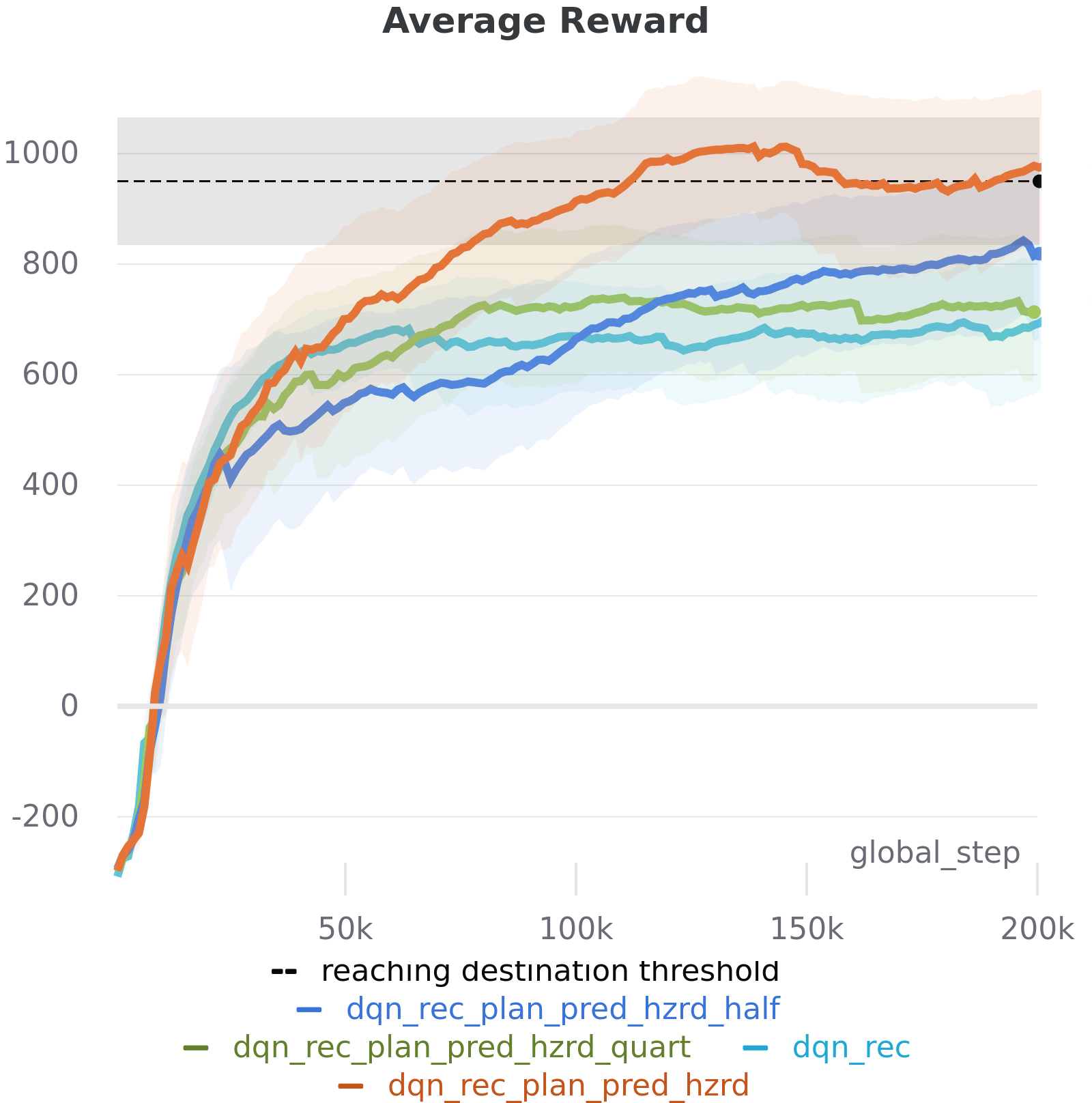}
         \caption{}
        %  \label{fig:five over x}
     \end{subfigure}
        \caption{Analysis of the effect of different dataset sizes on the down-stream RL task in two scenarios: (a) roundabout, (b) 5-way intersection.}
        \label{fig:dataset_size}
\end{figure}

The results, shown in Fig.~\ref{fig:dataset_size} for two scenarios, present better performance of our proposed method even when it was trained with a quarter of the dataset. The case trained with half of the dataset can solve the intersection scenario for some runs, but the model trained with a quarter of the dataset cannot; however, it outperformed the baseline.

\subsection{Qualitative Analysis}

\begin{figure}[t!]
\includegraphics[width=\columnwidth]{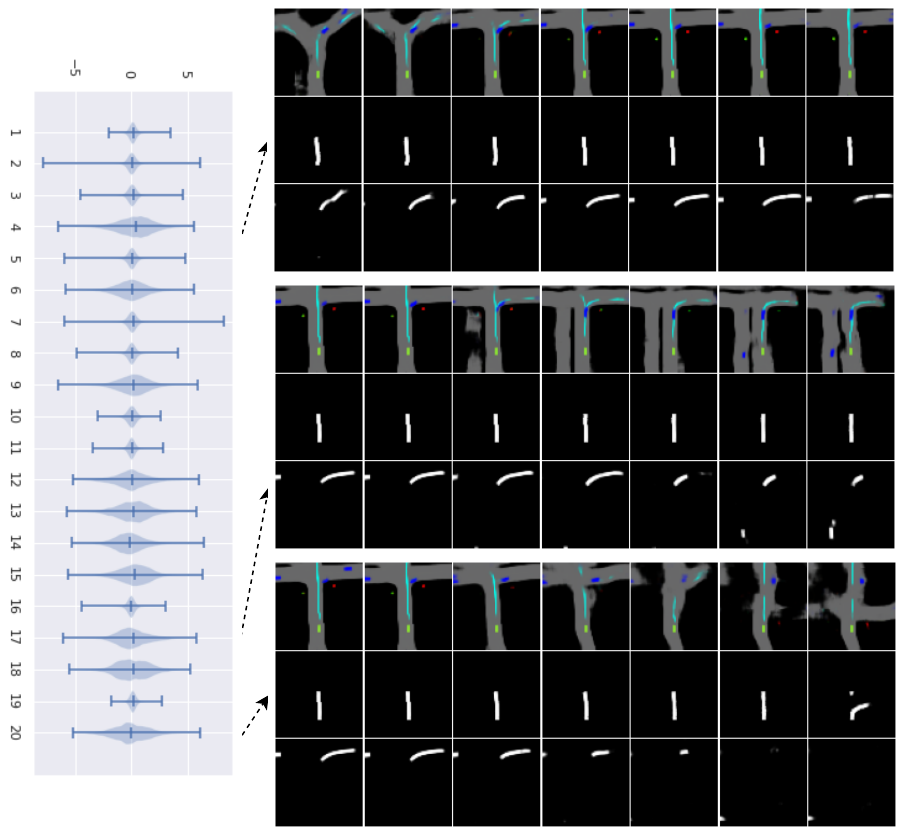}
\centering
\caption{Latent-space exploration. The left plot shows the violin-plot of the learned latent-space -- mean, standard deviation, min, and max -- for training dataset. The right hand side figures show outputs of the multi-head VAE when the \(i\)th latent element is changed from \( \mu_i - \sigma_i\) to \( \mu_i + \sigma_i\). \( \mu_i, \sigma_i\) are mean and standard deviation of the \(i\)th element for the whole training dataset. We show only the results of exploring three latent elements: 4th, 17th, and 20th. A video with more examples of latent space exploration can be found at: \href{https://youtu.be/5Tk8j6LXBmA}{\textit{https://youtu.be/5Tk8j6LXBmA}}.}
\label{fig:latent_viz}
\end{figure}

In this section, we explore the latent-space to answer question \ref{qual_analysis}. The learned latent-vector has 20 elements some of which have almost zero standard deviation and changing their value do not cause any difference in outputs.
We tested several latent space sizes. Low latent sizes push the network to mix information in latent elements. So, the disentanglement of latent elements will be lower, and by changing the latent element's value, multiple elements will change in the reconstruction heads. On the other hand, by considering higher latent space sizes, most of the elements will be zero, and it doesn't guarantee to have more disentanglement in latent space. We selected 20 as a number in between, but we have some latent elements close to zero with a low standard deviation. Note that we do not try to solve the disentanglement problem here.
Fig.~\ref{fig:latent_viz} shows the results for three latent elements. We used a sample input from dataset and encoded it to get the latent-vector. Then for each latent element \(i\), we changed its value from \( \mu_i - \sigma_i\) to \( \mu_i + \sigma_i\) while other latent elements were fixed. By exploring latent-space, we see changes in different factors in the scene such as road structure, traffic light color, route, dynamic objects pose, planning, and motion prediction. This shows the meaningfulness of the learned latent-space. We can also see the harmony between different heads' outputs when changing latent values.

\subsection{Generalization Analysis}

\begin{figure}[t!]
\includegraphics[width=0.4\columnwidth]{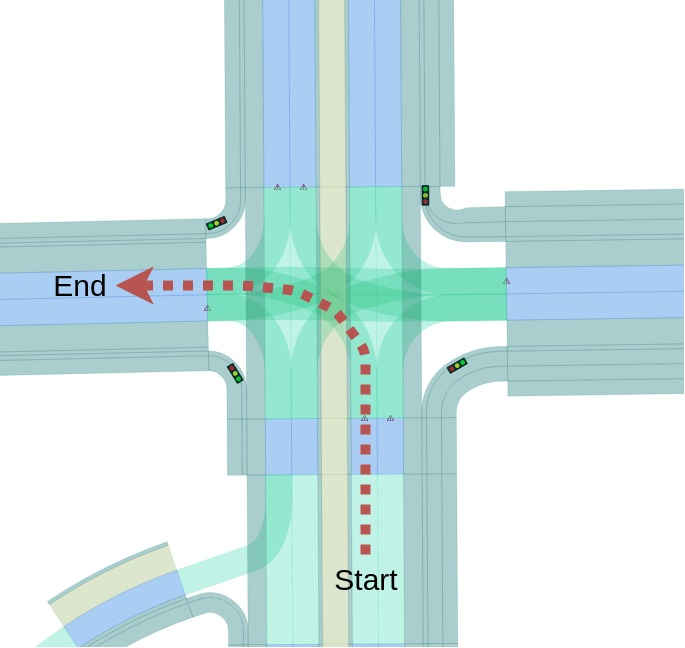}
\centering
\caption{\change{The 4-way intersection for generalization analysis.}}
\label{fig:generalization_scenario}
\end{figure}

\change{In order to answer question \ref{gnrlzn}, we evaluated the generalization capability of the proposed method by running the trained model on the 5-way intersection scenario in a new 4-way intersection scenario, shown in Fig~\ref{fig:generalization_scenario}, without any further training and fine-tuning.
}

\begin{table}[t!]
    \centering
  \caption{\change{The crash percentage, success rate, and rate of traffic light passing in 3 runs, 100 episodes each, for the trained model on the 5-way intersection scenario and evaluated in the 4-way intersection scenario.}}
  \label{tab:generalization}
  \begin{tabular}{|*{5}{c|}}\toprule
    \multirow{2}{*}{\textbf{\textit{Metric}}} & \multirow{2}{*}{\textbf{\textit{Method}}} & \multicolumn{3}{c|}{\textbf{\textit{Number of Cars}}} \\ 
     &  & \textit{10}     & \textit{50}     & \textit{100}    \\ \midrule
    
    \multirow{5}{*}{\textbf{\textit{Crash Percentage}}}
    
    & dqn\_cnn\_e2e  & 40 $\pm$ 2.8         & 45 $\pm$ 2.8               & 55 $\pm$ 2.5              \\
    & sac\_ae        & 31 $\pm$ 3.2        & 42 $\pm$ 2.1             & 53 $\pm$ 3.1            \\
    & curl           & 32 $\pm$ 3.5       & 39 $\pm$ 2.2             & 51 $\pm$ 2.6                \\
    & dqn\_rec       & 30 $\pm$ 2.6         & 38 $\pm$ 2.5             & 49 $\pm$ 3.9              \\
    & ours           & \textbf{20 $\pm$ 2.5} & \textbf{27 $\pm$ 2.1}  & \textbf{33 $\pm$ 3.6}   \\ \bottomrule

    \multirow{5}{*}{\textbf{\textit{Success Rate}}}
    
    & dqn\_cnn\_e2e  & 45 $\pm$ 2.3        & 40 $\pm$ 3.2               & 31 $\pm$ 3.1              \\
    & sac\_ae        & 50 $\pm$ 2.8        & 42 $\pm$ 2.6             & 34 $\pm$ 2.8            \\
    & curl           & 52 $\pm$ 3.2       & 43 $\pm$ 2.7             & 36 $\pm$ 3.4                \\
    & dqn\_rec       & 54 $\pm$ 2.9         & 48 $\pm$ 3.2             & 45 $\pm$ 3.1              \\
    & ours           & \textbf{72 $\pm$ 2.2} & \textbf{63 $\pm$ 2.1}  & \textbf{56 $\pm$ 2.4}   \\ \bottomrule
    
    \multirow{5}{*}{\textbf{\textit{Passing Traffic Light}}}
    
    & dqn\_cnn\_e2e  & -        & -             & 54 $\pm$ 2.5              \\
    & sac\_ae        & -        & -             & 38 $\pm$ 3.6            \\
    & curl           & -        & -             & 42 $\pm$ 2.9                \\
    & dqn\_rec       & -        & -             & 42 $\pm$ 3.2              \\
    & ours           & -        & -             & \textbf{26 $\pm$ 2.3}   \\ \bottomrule
    
  \end{tabular}
\end{table}

\change{
Due to no further training on the new scenario, the performance of all algorithms was lower than the 5-way intersection scenario in terms of crash percentage and success rate metrics. Nevertheless, our method outperformed other methods, as shown in Table~\ref{tab:generalization}.
Furthermore, we realized that sometimes the car sticks and doesn't move at all, which can be due to changing the scene and out of distribution input image.
The results also reveal that the passing traffic light metric does not change that much in the new 4-way intersection, which illustrates that the network has learned about the traffic light colors and rules and can generalize to new scenarios.
But the new road and scene structure causes weaker performance in the crash percentage and success rate.
}

% \addtolength{\textheight}{-3cm}
%%%%%%%%%%%%%%%%%%%%%%%%%%%%%%%%%%%%%%%%%%%%%%%%%%%%%%%%%%%%%%%%%%%%%%%%%%%%%%%%
\section{CONCLUSIONS}\label{conclusion}
Machine learning presents an important avenue to handle the complexity of situations in autonomous driving but its applicability is significantly hindered by the huge amounts of data needed for learning. 
In this paper, we proposed a multi-head VAE network with a bird's eye view input and task-relevant heads to learn efficient latent-space representations. 
These representations can then be used to train a driving policy more efficiently. 
Experimental comparison against baselines in two scenarios showed that the use of task-relevant heads in representation learning improves policy learning in four ways: the policy quality is better, the learning converges faster, policy learning requires less data, \change{and the learned policy generalizes better to new scenarios}. 
The proposed approach can be easily extended to other task-relevant factors if they can be encoded as images.

In real traffic environments, multiple vehicles with different driving policies interact with each other. 
In this case, generalization capability of a representation would likely benefit from disentanglement of that representation with respect to the different actors. 
The use of multi-task learning similar to this paper thus seems to provide a valuable avenue towards autonomous driving in complex traffic conditions. 

%%%%%%%%%%%%%%%%%%%%%%%%%%%%%%%%%%%%%%%%%%%%%%%%%%%%%%%%%%%%%%%%%%%%%%%%%%%%%%%%
% \appendices

% % you can choose not to have a title for an appendix
% % if you want by leaving the argument blank
% \section{}

%==========================================================

\section*{ACKNOWLEDGMENT}

The authors wish to acknowledge CSC – IT Center for Science, Finland, for generous computational resources.
We also acknowledge the computational resources provided by the Aalto Science-IT project.

%==========================================================

% \ifCLASSOPTIONcaptionsoff
%   \newpage
% \fi

% \addtolength{\textheight}{-10cm}

%==========================================================
\bibliographystyle{IEEEtran}
% \bibliography{{../Transactions-Bibliography/main}}
\bibliography{main}

%==========================================================
\begin{IEEEbiography}
[{
\includegraphics[width=1in,height=1.25in,clip,keepaspectratio]{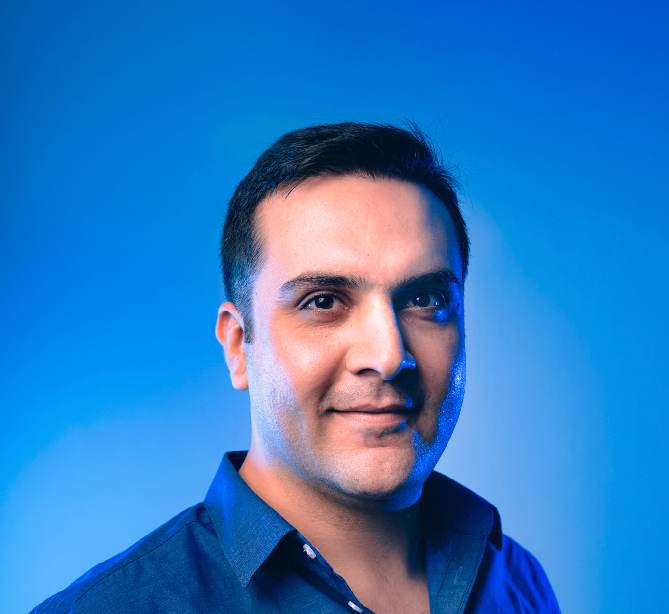}
}]
{Eshagh Kargar}
received the B.Sc. and M.Sc. degrees in electrical engineering from University of Tehran, Iran, in 2013 and 2016, respectively.
In 2019, he joined the Intelligent Robotics Group at Aalto University, Helsinki, Finland, where he is a Doctoral Candidate. His primary research interests include robotic perception, decision making, and learning.
\end{IEEEbiography}

\begin{IEEEbiography}
[{
\includegraphics[width=1in,height=1.25in,clip,keepaspectratio]{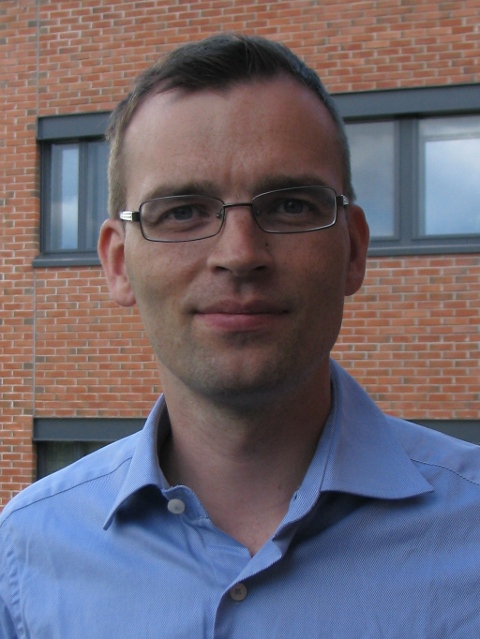}
}]
{Ville Kyrki}
(M'03–SM'13) received the M.Sc. and Ph.D. degrees in computer science from Lappeenranta University of Technology, Lappeenranta, Finland, in 1999 and 2002, respectively.

In 2003-2004, he was a Postdoctoral Fellow with the Royal Institute of Technology, Stockholm, Sweden, after which he returned to Lappeenranta University of Technology, holding various positions in 2003-2009. During 2009-2012, he was a Professor of computer science, Lappeenranta University of Technology. Since 2012, he is currently an Associate Professor of Intelligent Mobile Machines with Aalto University, Helsinki, Finland. His primary research interests include robotic perception, decision making, and learning.

Dr. Kyrki is a Fellow of Academy of Engineering Sciences (Finland), and a member of Finnish Robotics Society and Finnish Society of Automation. He was a Chair and Vice Chair of the IEEE Finland Section Jt. Chapter of CS, RA, and SMC Societies in 2012-2015 and 2015-2016, respectively, Treasurer of IEEE Finland Section in 2012-2013, and Co-Chair of IEEE RAS TC in Computer and Robot Vision in 2009-2013. He was an Associate Editor for the IEEE Transactions on Robotics in 2014-2017.
\end{IEEEbiography}

%\newpage



%\vfill


\end{document}